\PassOptionsToPackage{unicode}{hyperref}
\PassOptionsToPackage{hyphens}{url}
\documentclass[
]{article}
\usepackage{amsmath,amssymb}
\usepackage{iftex}
\ifPDFTeX
  \usepackage[T1]{fontenc}
  \usepackage[utf8]{inputenc}
  \usepackage{textcomp} 
\else 
  \usepackage{unicode-math} 
  \defaultfontfeatures{Scale=MatchLowercase}
  \defaultfontfeatures[\rmfamily]{Ligatures=TeX,Scale=1}
\fi
\usepackage{lmodern}
\ifPDFTeX\else
\fi
\IfFileExists{upquote.sty}{\usepackage{upquote}}{}
\IfFileExists{microtype.sty}{
  \usepackage[]{microtype}
  \UseMicrotypeSet[protrusion]{basicmath} 
}{}
\makeatletter
\@ifundefined{KOMAClassName}{
  \IfFileExists{parskip.sty}{%
    \usepackage{parskip}
  }{
    \setlength{\parindent}{0pt}
    \setlength{\parskip}{6pt plus 2pt minus 1pt}}
}{
  \KOMAoptions{parskip=half}}
\makeatother
\usepackage{xcolor}
\usepackage{longtable,booktabs,array}
\usepackage{calc} 
\usepackage{etoolbox}
\makeatletter
\patchcmd\longtable{\par}{\if@noskipsec\mbox{}\fi\par}{}{}
\makeatother
\IfFileExists{footnotehyper.sty}{\usepackage{footnotehyper}}{\usepackage{footnote}}
\makesavenoteenv{longtable}
\usepackage{graphicx}
\makeatletter
\def\maxwidth{\ifdim\Gin@nat@width>\linewidth\linewidth\else\Gin@nat@width\fi}
\def\maxheight{\ifdim\Gin@nat@height>\textheight\textheight\else\Gin@nat@height\fi}
\makeatother
\setkeys{Gin}{width=\maxwidth,height=\maxheight,keepaspectratio}
\makeatletter
\def\fps@figure{htbp}
\makeatother
\ifLuaTeX
  \usepackage{selnolig}  
\fi
\usepackage{bookmark}
\IfFileExists{xurl.sty}{\usepackage{xurl}}{} 
\hypersetup{
  hidelinks,
  pdfcreator={LaTeX via pandoc}}

\author{}
\date{}

\begin{document}

\textbf{SETAREH Transform: A Sorting-Assisted Haar-Wavelet\\
Sparsification Method for Approximate\\
Inner-Product Computation}

\emph{Seyed Kamaledin Setarehdan}

\emph{Control and Intelligent Processing Centre of Excellence, School of
Electrical and Computer Engineering, College of Engineering, University
of Tehran, Tehran, Iran}

\section{Abstract}\label{abstract}

Modern deep learning systems increasingly rely on high-dimensional vector operations, in which large input and parameter vectors are repeatedly stored, transferred, and processed. Although substantial research has focused on reducing neural-network model size through pruning, quantization, knowledge distillation, and efficient architectures, the storage, memory-traffic, and communication costs associated with repeatedly presented high-dimensional input vectors remain important practical considerations, particularly in distributed and edge-computing environments.

This paper proposes the SETAREH (Sorting-assisted Energy-concentrating Transform for Approximate Representation using Haar wavelets) transform, a Haar-wavelet-based sparsification approach for approximating the inner product between a fixed high-dimensional coefficient vector and successive input vectors. By applying the same sequence of permutations and Haar transforms to both vectors, SETAREH preserves the exact inner product before truncation. It then retains only a subset of the transformed coefficient-input pairs, reducing their representation cost while introducing a controllable approximation error.

To evaluate SETAREH, we consider two complementary experimental settings: a synthetic Gaussian-vector baseline and a real apparent-age estimation task using a convolutional neural network on the APPA-REAL dataset.

The results show that SETAREH can approximate the evaluated inner products using substantially fewer transformed coefficient-input pairs than the original vector dimension. The findings support its use as a potential structured data-representation method for repeated inner products with fixed weights, while the practical benefit depends on approximation accuracy and the cost of storing or transmitting permutation metadata.

Keywords: deep learning, data reduction, sparse representation, inner
product, Haar wavelet transform, discrete wavelet transform, vector
compression, edge computing, bandwidth reduction, neural networks, VGG16

\section{1. Introduction}\label{introduction}

The rapid development and scaling of deep learning have resulted in
neural network models with increasingly large numbers of parameters,
wider intermediate representations, and higher-dimensional activation
vectors {[}1--3{]}. The increasing size and computational complexity of
deep neural networks have consequently motivated extensive research into
efficient architectures. Modern architectures range from convolutional
neural networks to attention-based and Transformer architectures, with
increasing model capacity serving as an important mechanism for
improving predictive performance. Representative architectures such as
ResNet, MobileNet, and EfficientNet illustrate different approaches to
balancing predictive accuracy, computational complexity, memory
requirements, and model size {[}4--6{]}. Residual networks demonstrated
that very deep models could be effectively optimized through residual
learning while maintaining practical computational requirements {[}4{]}.
MobileNets introduced depthwise separable convolutions as a mechanism
for substantially reducing computation and model size, particularly for
mobile and embedded applications {[}5{]}. EfficientNet further
demonstrated that jointly scaling network depth, width, and input
resolution can achieve improved accuracy--efficiency trade-offs compared
with conventional scaling strategies {[}6{]}.

At the same time, the computational burden of deep neural networks is
not limited to arithmetic operations. Memory access, storage, and data
movement can represent significant system-level costs, particularly when
models and intermediate data must be processed on resource-constrained
devices or transferred between computational units. Consequently,
neural-network efficiency has been studied from several complementary
perspectives, including architectural design, parameter pruning,
quantization, knowledge distillation, and model compression {[}4--9{]}.
For example, Han et al. proposed Deep Compression, which combines
pruning, trained quantization, and Huffman coding to substantially
reduce neural-network storage requirements. Their work demonstrated that
large reductions in model storage can be achieved while preserving
predictive accuracy {[}7{]}. Efficient architecture design has similarly
been investigated in MobileNets and EfficientNets, where reductions in
computation and model complexity are obtained through architectural
choices and scaling strategies {[}5,6{]}.

Beyond model compression and architectural optimization, reducing data
movement has also emerged as an important consideration in efficient
neural-network processing. Chen et al. demonstrated that data movement
and memory access can represent major sources of energy consumption in
deep neural-network processing. Their Eyeriss architecture explicitly
targets data movement through the memory hierarchy by exploiting data
reuse and an optimized dataflow, highlighting the importance of reducing
data movement in addition to reducing arithmetic operations {[}8{]}.
Quantization provides another widely studied approach to reducing the
memory footprint and computational cost of neural networks. Jacob et al.
proposed an integer-only inference framework that quantizes both weights
and activations to lower-bit representations while maintaining accuracy
close to floating-point inference {[}9{]}. Most existing approaches,
however, primarily focus on reducing model size, computational
complexity, or the representation cost of model parameters and
activations.

A different but related problem arises when a fixed coefficient vector
is repeatedly applied to a large number of input vectors to compute
their inner products. Such inner-product operations arise in many
neural-network computations, including fully connected layers and, more
generally, linear projections used in convolutional and attention-based
architectures. In these settings, the same coefficient vector may
participate in a large number of successive inner-product operations,
while the corresponding input vectors repeatedly enter the computation.
Consequently, reducing the amount of input data that must be stored,
transferred, or processed for each inner-product operation could provide
a complementary form of data reduction without modifying the underlying
network architecture or its learned parameters.

This observation motivates the approach proposed in this work. Rather
than compressing the neural-network model itself, we investigate whether
the permutation invariance of the inner product can be exploited to
construct a compact representation of repeatedly occurring
high-dimensional input vectors. By appropriately reordering a fixed
coefficient vector, applying a Haar Discrete Wavelet Transform, and
promoting a more compact distribution of its transform-domain
representation, the proposed method seeks to approximate the original
inner product using only a small subset of the corresponding input
samples. We refer to the proposed framework as SETAREH (Sorting-assisted
Energy-concentrating Transform for Approximate Representation using Haar
wavelets). The method is designed as a complementary data-reduction
technique, with potential applications in memory-constrained processing,
repeated inference, and communication- or bandwidth-constrained
neural-network systems.

In this work, two variants of the proposed framework are investigated.
SETAREH-3 applies an initial sorting operation, followed by a Haar
Discrete Wavelet Transform and a second sorting operation. SETAREH-5
extends this procedure by repeating the sorting and Haar-transform
stages once before the final sorting operation. The two variants
therefore provide different trade-offs between approximation accuracy,
representation compactness, and computational cost. Their performance is
systematically evaluated across different vector dimensions and retained
coefficient counts, with particular emphasis on approximation error and
the number of retained input samples.

The remainder of the paper is organized as follows. Section 2 introduces
the proposed SETAREH framework and formally describes the SETAREH-3 and
SETAREH-5 processing procedures. Section 3 presents the experimental
methodology and defines the two experimental settings, statistical
evaluation protocol, and performance metrics. Section 4 reports and
analyses the results, including approximation accuracy, robustness,
compression characteristics of SETAREH and also the effects of
vector-dimension on its performance using synthetic and real scenarios.
Finally, section 5 discusses the implications, limitations, and
potential applications of the proposed algorithm.

\section{2. SETAREH Transform
Framework}\label{setareh-transform-framework}

In this section we first introduce and formulate the proposed SETAREH
transform framework and formally describe its two variants, namely
SETAREH-3 and SETAREH-5.

Let \textbf{W} and \textbf{X} denote a fixed coefficient vector and an
input vector, respectively, both of length \emph{N.} The inner product
of the two vectors, S, is calculated as shown in Eq. (1).

\(\mathbf{W},\mathbf{X}\in\mathbb{R}^{N},\quad S=\mathbf{W}^{T}\mathbf{X}=\sum_{i=1}^{N}w_i x_i.\) \hfill (1)

The objective is to approximate the conventional inner product of Eq.
(1) using substantially fewer than \emph{N} coefficient--input pairs.
The proposed approach is specifically intended for the setting in which
the coefficient vector \textbf{W} is fixed or changes relatively
infrequently, whereas a potentially large number of input vectors
\textbf{X} are subsequently presented to the computation. This setting
is particularly relevant to repeated applications of the same linear
operation within the neural networks.

The central property exploited by SETAREH is the permutation invariance
of the inner product. For any \emph{N×N} permutation matrix \textbf{P},
the same permutation can be applied to both vectors without changing
their exact inner product as shown in Eq. (2).

\[
(\mathbf{P}\mathbf{W})^{\mathsf T}(\mathbf{P}\mathbf{X})
=\mathbf{W}^{\mathsf T}\mathbf{P}^{\mathsf T}\mathbf{P}\mathbf{X}
=\mathbf{W}^{\mathsf T}\mathbf{X}.
\tag{2}
\]

Now, consider a permutation of the coefficient vector in which
similar-valued elements are placed close to one another. Such a
permutation can be obtained by sorting \textbf{W} in ascending order.
Using the same permutation matrix, the associated input vector
\textbf{X} is also reordered as shown in (3):

\[
\mathbf{W}^{(1)}=\mathbf{P}_1\mathbf{W},\qquad
\mathbf{X}^{(1)}=\mathbf{P}_1\mathbf{X}.
\tag{3}
\]

Therefore;

\[
(\mathbf{W}^{(1)})^{\mathsf T}\mathbf{X}^{(1)}
=(\mathbf{P}_1\mathbf{W})^{\mathsf T}\mathbf{P}_1\mathbf{X}
=\mathbf{W}^{\mathsf T}\mathbf{X}.
\tag{4}
\]

Sorting places similar coefficient values next to each other and
therefore reduces abrupt local variations in the reordered coefficient
sequence. This property motivates the subsequent application of the Haar
Discrete Wavelet Transform (HDWT). In its simplest form, HDWT generates
approximation coefficients from local averages and detail coefficients
from local differences between neighboring samples. Consequently, abrupt
variations between adjacent elements tend to produce relatively large
detail coefficients, whereas neighboring values with similar magnitudes
produce smaller detail coefficients. The ordering of the elements of a
vector can therefore influence the distribution of its energy across the
wavelet coefficients. In this sense, the sorting operation acts as a
wavelet-oriented preprocessing or preconditioning step.

Let H denote the orthonormal Haar wavelet transform matrix {[}10-12{]}.
The transformed coefficient and input vectors are obtained by applying
the same transform to both vectors as shown in Eq. (5). Since the Haar transform matrix is orthonormal, it satisfies\(\mathbf{H}^{\mathsf T}\mathbf{H}=\mathbf{I}\). Therefore, applying the same transform to both vectors preserves the exact inner product, as shown in Eq. (6).

\[
\mathbf{W}^{(2)}=\mathbf{H}\mathbf{W}^{(1)},\qquad
\mathbf{X}^{(2)}=\mathbf{H}\mathbf{X}^{(1)}.
\tag{5}
\]

\[
\begin{aligned}
(\mathbf{W}^{(2)})^{\mathsf T}\mathbf{X}^{(2)}
&=(\mathbf{H}\mathbf{W}^{(1)})^{\mathsf T}
  (\mathbf{H}\mathbf{X}^{(1)})\\
&=(\mathbf{W}^{(1)})^{\mathsf T}\mathbf{H}^{\mathsf T}\mathbf{H}\mathbf{X}^{(1)}
=(\mathbf{W}^{(1)})^{\mathsf T}\mathbf{X}^{(1)}
=\mathbf{W}^{\mathsf T}\mathbf{X}.
\end{aligned}
\tag{6}
\]

Consequently, the transformation procedure including the sorting
operator and the HDWT does not introduce approximation error to the
inner product. Approximation is introduced only when the transformed
vectors are truncated.

To provide a unified description of the proposed variants of SETAREH,
let \textbf{P}\emph{\textsubscript{j}} denote a permutation matrix
obtained by sorting the current coefficient vector at stage \emph{j}.
Each sorting or Haar-transform operation is applied identically to the
coefficient vector and its associated input vector. Therefore, for the
permuted fixed coefficient vector $\mathbf{W}^{(1)}$ and the permuted input
vector $\mathbf{X}^{(1)}$ of Equation (3), the SETAREH-3 and SETAREH-5 are
defined as follows:

\[
\begin{aligned}
\text{SETAREH-3:}\quad &\mathbf{W}^{(3)}=\mathbf{P}_2\mathbf{H}\mathbf{W}^{(1)},\\
&\mathbf{X}^{(3)}=\mathbf{P}_2\mathbf{H}\mathbf{X}^{(1)}.
\end{aligned}
\tag{7}
\]

\[
\begin{aligned}
\text{SETAREH-5:}\quad &\mathbf{W}^{(5)}=\mathbf{P}_3\mathbf{H}\mathbf{P}_2\mathbf{H}\mathbf{W}^{(1)},\\
&\mathbf{X}^{(5)}=\mathbf{P}_3\mathbf{H}\mathbf{P}_2\mathbf{H}\mathbf{X}^{(1)}.
\end{aligned}
\tag{8}
\]

Since every permutation matrix and every Haar transform is orthonormal,
the complete SETAREH transformation preserves the exact inner product.
It is important to notice that SETAREH does not approximate the inner
product through the transformation itself; approximation occurs through
coefficient truncation in the following steps.

\subsection{2.1 Approximate Inner-Product Computation and Performance
Measures}\label{approximate-inner-product-computation-and-performance-measures}

Using the vector pairs obtained by SETAREH-3 and SETAREH-5, the exact
inner product remains equal to the original value before truncation.
After transformation and final sorting, only the first \emph{K} paired coefficients are retained, producing the approximate inner products \(\widehat{S}_{3,K}\) and \(\widehat{S}_{5,K}\), as shown in Eq. (9).

\[
\widehat{S}_{3,K}=\sum_{i=1}^{K}w_i^{(3)}x_i^{(3)},\qquad
\widehat{S}_{5,K}=\sum_{i=1}^{K}w_i^{(5)}x_i^{(5)}.
\tag{9}
\]

To quantify the approximation performance of SETAREH, we conducted
statistical analyses in two complementary experimental settings,
described in the following sections. To provide a scale-independent
measure of error, the Relative Errors are defined as:

\[
\mathrm{RE}_{3,K}=\frac{|S-\widehat{S}_{3,K}|}{|S|}\times100\%,\qquad
\mathrm{RE}_{5,K}=\frac{|S-\widehat{S}_{5,K}|}{|S|}\times100\%.
\tag{10}
\]

Here, \(S\) is the exact inner product, and \(\widehat{S}_{3,K}\) and \(\widehat{S}_{5,K}\) are the corresponding truncated approximations defined in Eqs. (1) and (9).
Because the magnitude \(|S|\) of the exact inner product can occasionally be close to zero, the relative errors defined in Eq. (10) can become disproportionately large even when the absolute error is small.

To avoid domination of the relative-error statistics by this
near-zero-reference regime, we conditioned the analysis on the magnitude
of the exact inner product. Specifically, the lowest 10\% of trials
ranked according to \textbar S\textbar{} were excluded from the
relative-error aggregation. Then, the median value of the remaining
errors is identified and used as the \emph{Conditioned Median Relative
Error} (\emph{CMRE}) as the principal summary comparative measure in
this study. If the approximate inner product of Eq. (9) produces small
enough error causing no inference error by the neural network, then only
the \emph{K} coefficients need to be stored or transmitted instead of
the original large number of \emph{N} coefficients.

A central feature of SETAREH is the possibility to break the whole
process into offline processing of the fixed transformed coefficient vector
$\mathbf{W}^{(1)}$ and online processing of subsequently presented input vectors
$\mathbf{X}^{(1)}$, as schematically shown in Figure 1. For a fixed coefficient
vector, the required sorting and Haar-transform operations can be
performed once offline. The resulting transformed coefficient vector and
associated permutation information are then retained for subsequent
online computations on each new input vector. For SETAREH-3, the online
transformation requires two permutation operations and one Haar
transform, whereas SETAREH-5 requires three permutation operations and
two Haar transforms. The two variants therefore introduce different
computational requirements in exchange for potentially different
approximation accuracies and smaller number of coefficients to be stored
or transmitted. In continue the experimental methodology to evaluate the
data reduction performance of SETAREH on synthetic data and also on
examples of real-world neural network applications is presented.

\includegraphics[width=4.92736in,height=3.3913in]{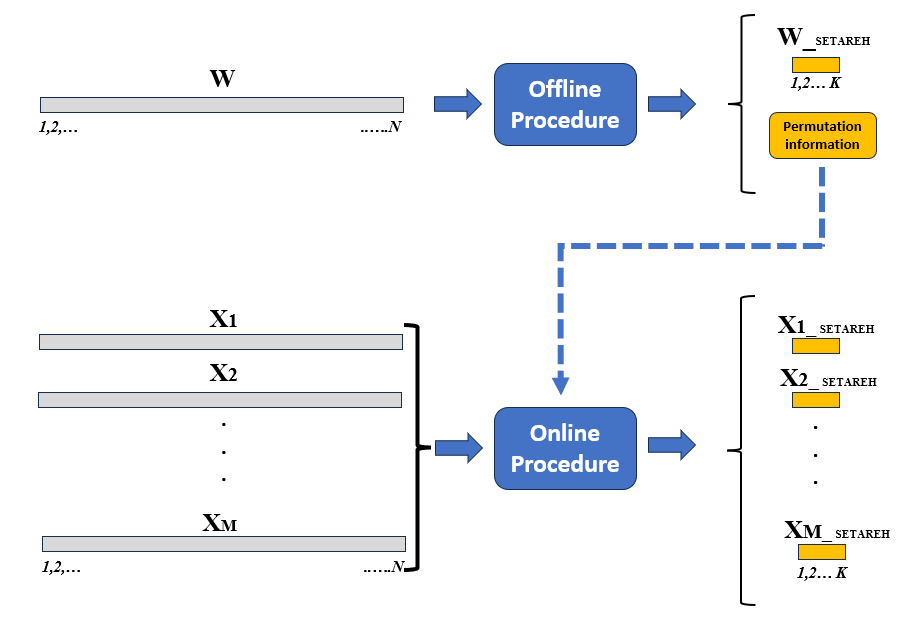}

Figure 1) Schematic description of the SETAREH transform framework.
Coefficients to be stored or transferred needed for each inner-product
operation before SETAREH transform (gray) and after that (orange).

\section{}\label{section}

\section{3. Experimental Methodology}\label{experimental-methodology}

\subsection{3.1 Experimental Settings}\label{experimental-settings}

To evaluate the SETAREH framework two experiments based on synthetic and
real scenarios are considered. The first experiment uses independently
generated Gaussian random vectors as coefficient and input vectors and
serves as a controlled numerical baseline for analyzing and comparing
the behavior and robustness of SETAREH-3 and SETAREH-5 across different
vector dimensions and various retention levels. The vector dimensions
$N \in \{1024, 2048, 4096, 8192, 16384, 65536\}$ were selected from
powers of two so that the full dyadic Haar transform could be applied
without boundary extension. For each of the 50 different coefficient
vectors, 50 independent input vectors were also generated providing a
total of 2500 inner products. The same coefficient vectors and exactly
the same input vectors were used for SETAREH-3 and SETAREH-5, ensuring a
paired comparison.

The second experiment replaces the randomly generated coefficient
vectors with the actual 4096-dimensional regression vector derived using
the training image set and the 4096-dimensional feature vectors
extracted from the fc2 layer of a pretrained VGG16 network {[}13{]} on
the validation and test image sets of the APPA-REAL dataset {[}20{]} for
apparent age estimation.

\subsection{\texorpdfstring{3.2 Data Reduction and Compression Metrics
}{3.2 Data Reduction and Compression Metrics }}\label{data-reduction-and-compression-metrics}

To measure the reduction in the number of values participating in the
approximate inner-product computation using the SETAREH algorithm, the
number of retained coefficient--input pairs, \emph{K}, compared with the
original vector dimension \emph{N}. A coefficient-only \emph{Compression
Ratio} is therefore defined as:

\begin{equation}\label{eq:cr-coeff}\mathrm{CR}^{*}=\frac{N}{K}\tag{11}\end{equation}

It must be noted that, \emph{CR*} represents
\emph{coefficient/input-value} reduction only and does not include the
storage required for permutation information. Therefore, it is not
interpreted directly as the total model-storage compression ratio.

For a comparative robustness analysis of the SETAREH variants, three
complementary perspectives are also considered. First, for fixed
\emph{N} and \emph{K}, the error-reduction factor is evaluated as the
ratio of SETAREH-3 error to SETAREH-5 error as shown by
\emph{G\textsubscript{K}} in Eq. (12). Second, for a specified target
conditioned median relative error (\emph{CMRE\textsubscript{target}}),
the minimum retained coefficient counts \emph{K\textsubscript{min,3
}}and \emph{K\textsubscript{min,5 }} are determined. Third, the
corresponding coefficient-only compression ratio at the target accuracy
(\emph{CR\textsubscript{3}*} and \emph{CR\textsubscript{5}*}) are
calculated as defined below.

\[G_K = \frac{\mathrm{CMRE}_{3,K}}{\mathrm{CMRE}_{5,K}}.\tag{12}\]

\[K_{\min,3}=\min\{K:\mathrm{CMRE}_{3,K}\leq\mathrm{CMRE}_{\mathrm{target}}\},\qquad K_{\min,5}=\min\{K:\mathrm{CMRE}_{5,K}\leq\mathrm{CMRE}_{\mathrm{target}}\}.\tag{13}\]

\[\mathrm{CR}_3^*=\frac{N}{K_{\min,3}},\qquad \mathrm{CR}_5^*=\frac{N}{K_{\min,5}}.\tag{14}\]

\section{4. Results and Discussion}\label{results-and-discussion}

As explained in section 3.1, to evaluate the SETAREH framework, two
experimental settings based on synthetic and real scenarios considered
in this study. The results for each experiment are presented below.

\textbf{4.1 The Results of the Synthetic Baseline}

The first experimental setting was designed as a controlled numerical
baseline using independently generated Gaussian coefficient and input
vectors. The approximation performance of SETAREH-3 and SETAREH-5 was
evaluated as a function of the number of retained coefficient--input
pairs \emph{K} across the six investigated vector dimensions.
\emph{CMRE} was selected as the primary comparison metric because it
provides a robust measure of typical approximation accuracy while
reducing the influence of near-zero exact inner products. Figure 2 and
Table 1 demonstrate the results of the whole experimental study carried
out for SETAREH-3 and SETAREH-5. As seen, across all investigated
dimensions, a clear reduction in approximation error was observed as
\emph{K} increased for both SETAREH variants.

More importantly, the reduction achieved by SETAREH-5 became
substantially greater than that of SETAREH-3 as the retained coefficient
count increased. As a comparative parameter, the error-reduction factor
in Table 1 is calculated as ($\mathrm{CMRE}_{3,K}$
\textbf{(\%) /} $\mathrm{CMRE}_{5,K}$
\textbf{(\%)}). Values greater than one indicates lower error for
SETAREH-5. These results demonstrate that the additional transformation
stage in SETAREH-5 can substantially improve approximation accuracy when
a moderate number of transformed coefficient pairs are retained.

For \emph{N}=4096, for example, the conditioned median relative error
decreased from 14.11\% at \emph{K}=10 to 2.98\% at \emph{K}=50 for
SETAREH-3. Under the same conditions, SETAREH-5 reduced the error to
only 0.272\%. The difference became increasingly pronounced at larger
values of \emph{K}. At \emph{K}=100, the conditioned median relative
errors were 1.59\% and 0.0623\% for SETAREH-3 and SETAREH-5,
respectively and at \emph{K}=200, these values decreased to 0.863\% and
0.0141\%. Although all experiments were carried out for all six
predefined vector sizes, for article length considerations only the
results for the vector size \emph{N}= 4096 will be illustrated in
continue. These results establish the numerical baseline against which
the subsequent experiments using real neural network scenarios can be
compared.

\includegraphics[width=6.8in,height=3.40764in]{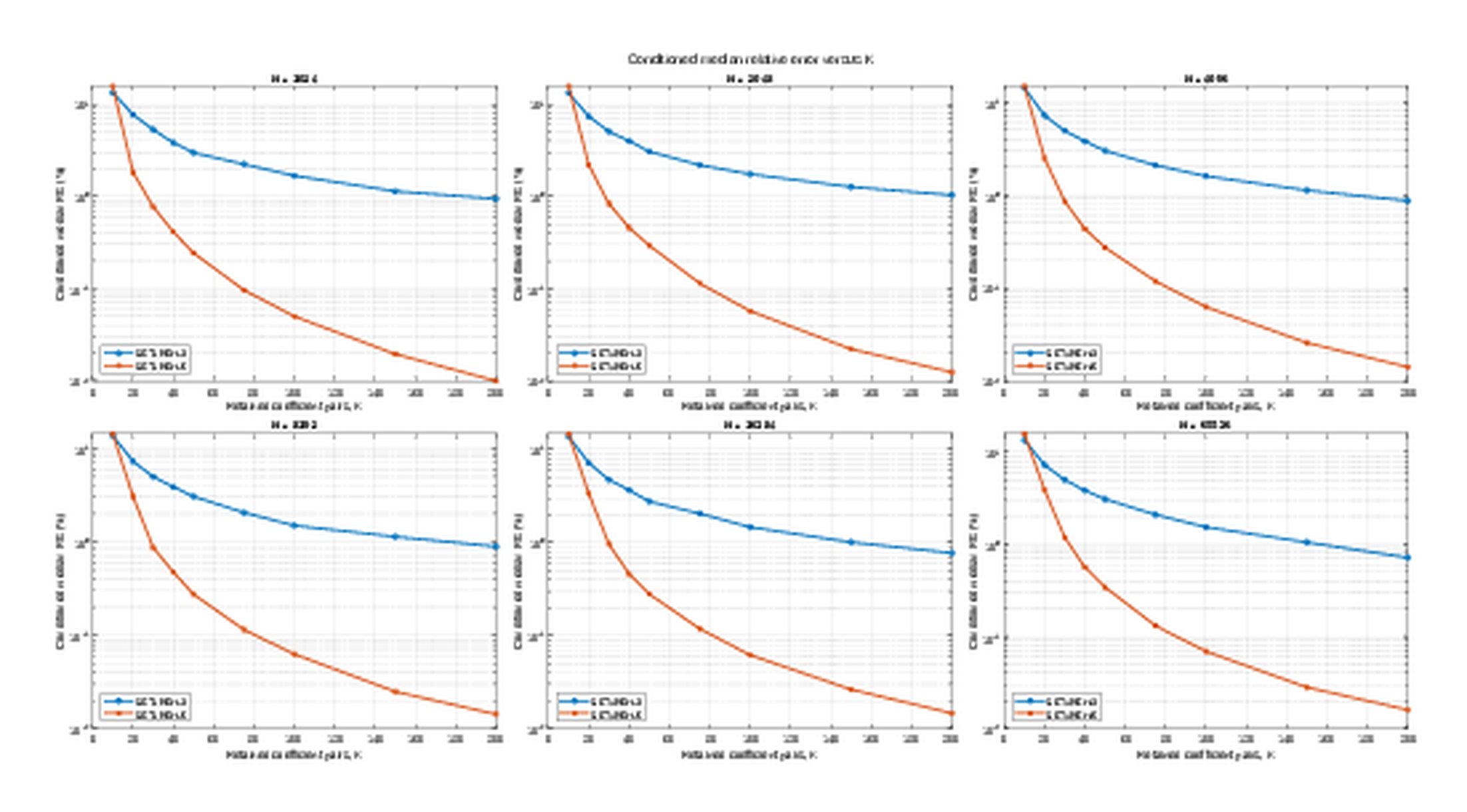}

Figure 2) Conditioned median relative error versus \emph{K} for
SETAREH-3 (in blue) and SETAREH-5 (in red) across the six investigated
vector dimensions of the synthetic baseline data set.

Table 1) Conditioned median relative error (\%) for the synthetic
baseline

\begin{longtable}[]{@{}
  >{\raggedright\arraybackslash}p{(\columnwidth - 8\tabcolsep) * \real{0.1092}}
  >{\raggedright\arraybackslash}p{(\columnwidth - 8\tabcolsep) * \real{0.0875}}
  >{\raggedright\arraybackslash}p{(\columnwidth - 8\tabcolsep) * \real{0.2187}}
  >{\raggedright\arraybackslash}p{(\columnwidth - 8\tabcolsep) * \real{0.2723}}
  >{\raggedright\arraybackslash}p{(\columnwidth - 8\tabcolsep) * \real{0.3124}}@{}}
\toprule\noalign{}
\begin{minipage}[b]{\linewidth}\raggedright
\emph{N}
\end{minipage} & \begin{minipage}[b]{\linewidth}\raggedright
\emph{K}
\end{minipage} & \begin{minipage}[b]{\linewidth}\raggedright
$\mathrm{CMRE}_{3,K}$ (\%)
\end{minipage} & \begin{minipage}[b]{\linewidth}\raggedright
$\mathrm{CMRE}_{5,K}$ (\%)
\end{minipage} & \begin{minipage}[b]{\linewidth}\raggedright
Error reduction

$G_K$
\end{minipage} \\
\midrule\noalign{}
\endhead
\bottomrule\noalign{}
\endlastfoot
1024 & 10 & 13.3692 & 15.6898 & 0.85 \\
1024 & 20 & 7.7074 & 1.8232 & 4.23 \\
1024 & 30 & 5.2886 & 0.7715 & 6.85 \\
1024 & 40 & 3.8110 & 0.4117 & 9.26 \\
1024 & 50 & 2.9607 & 0.2426 & 12.21 \\
1024 & 75 & 2.2286 & 0.0959 & 23.24 \\
1024 & 100 & 1.6695 & 0.0498 & \textbf{33.52} \\
1024 & 150 & 1.1341 & 0.0194 & 58.52 \\
1024 & 200 & 0.9378 & 0.0099 & \textbf{94.35} \\
2048 & 10 & 13.2646 & 15.7167 & 0.84 \\
2048 & 20 & 7.3667 & 2.1917 & 3.36 \\
2048 & 30 & 5.0429 & 0.8345 & 6.04 \\
2048 & 40 & 3.9864 & 0.4582 & 8.70 \\
2048 & 50 & 3.0635 & 0.2935 & 10.44 \\
2048 & 75 & 2.1900 & 0.1141 & 19.19 \\
2048 & 100 & 1.7418 & 0.0572 & \textbf{30.47} \\
2048 & 150 & 1.2668 & 0.0221 & 57.34 \\
2048 & 200 & 1.0350 & 0.0124 & \textbf{83.36} \\
\textbf{4096} & \textbf{10} & \textbf{14.1115} & \textbf{14.7985} &
0.95 \\
\textbf{4096} & 20 & 7.1397 & 2.4727 & 2.89 \\
\textbf{4096} & 30 & 4.8900 & 0.8510 & 5.75 \\
\textbf{4096} & 40 & 3.7662 & 0.4301 & 8.76 \\
\textbf{4096} & \textbf{50} & \textbf{2.9838} & \textbf{0.2718} &
\textbf{10.98} \\
\textbf{4096} & 75 & 2.0812 & 0.1177 & 17.68 \\
\textbf{4096} & \textbf{100} & \textbf{1.5902} & \textbf{0.0623} &
\textbf{25.52} \\
\textbf{4096} & 150 & 1.1192 & 0.0257 & 43.56 \\
\textbf{4096} & \textbf{200} & \textbf{0.8633} & \textbf{0.0141} &
\textbf{61.22} \\
8192 & 10 & 13.6802 & 14.3182 & 0.96 \\
8192 & 20 & 7.2635 & 3.0659 & 2.37 \\
8192 & 30 & 4.9773 & 0.8707 & 5.72 \\
8192 & 40 & 3.8703 & 0.4744 & 8.16 \\
8192 & 50 & 3.0553 & 0.2738 & 11.16 \\
8192 & 75 & 2.0534 & 0.1144 & 17.95 \\
8192 & 100 & 1.4896 & 0.0626 & \textbf{23.78} \\
8192 & 150 & 1.1336 & 0.0249 & 45.62 \\
8192 & 200 & 0.8894 & 0.0142 & \textbf{62.56} \\
16384 & 10 & 13.7158 & 14.7137 & 0.93 \\
16384 & 20 & 7.1626 & 3.3715 & 2.12 \\
16384 & 30 & 4.7167 & 0.9580 & 4.92 \\
16384 & 40 & 3.6482 & 0.4585 & 7.96 \\
16384 & 50 & 2.7548 & 0.2791 & 9.87 \\
16384 & 75 & 2.0441 & 0.1181 & 17.32 \\
16384 & 100 & 1.4609 & 0.0616 & \textbf{23.70} \\
16384 & 150 & 1.0059 & 0.0263 & 38.23 \\
16384 & 200 & 0.7707 & 0.0146 & \textbf{52.71} \\
65536 & 10 & 13.5615 & 15.8494 & 0.86 \\
65536 & 20 & 7.2636 & 3.9097 & 1.86 \\
65536 & 30 & 5.0045 & 1.1870 & 4.22 \\
65536 & 40 & 3.8412 & 0.5630 & 6.82 \\
65536 & 50 & 3.0909 & 0.3427 & 9.02 \\
65536 & 75 & 2.0984 & 0.1308 & 16.05 \\
65536 & 100 & 1.5370 & 0.0689 & \textbf{22.30} \\
65536 & 150 & 1.0524 & 0.0280 & 37.58 \\
65536 & 200 & 0.7176 & 0.0160 & \textbf{44.89} \\
\end{longtable}

As Table 1 shows, the results at \emph{K}=10 reveal a distinct operating
regime for all vector sizes. Unlike the results at larger values of
\emph{K}, SETAREH-5 does not consistently outperform SETAREH-3 when only
ten coefficient--input pairs are retained. For example, at
\emph{N}=4096, the conditioned median relative errors are 14.11\% for
SETAREH-3 and 14.80\% for SETAREH-5. At such extreme retention levels,
the approximation depends on a very small number of transformed
coefficient--input pairs and is therefore highly sensitive to the
detailed distribution and signs of the retained coefficients. As
\emph{K} increases, this sensitivity decreases and the advantage of
SETAREH-5 becomes increasingly consistent. The \emph{K}=10 results are
therefore useful as a stress test of the extreme-compression regime
rather than as the representative operating point of the method.

\subsection{4.1.1 Accuracy Improvement Achieved by
SETAREH-5}\label{accuracy-improvement-achieved-by-setareh-5}

The accuracy advantage of SETAREH-5 in comparison to SETAREH-3 can be
quantified directly through the ratio between the corresponding errors.
The improvement is relatively modest in the extreme \emph{K}=10 regime
but increases rapidly with \emph{K}. As shown in Table 1, the accuracy
advantage of SETAREH-5 becomes increasingly pronounced as the number of
retained coefficients increases. For example, at \emph{N}=4096,
SETAREH-5 reduces the conditioned median relative error by approximately
11-fold, 25-fold, and 61-fold at \emph{K}=50, 100, and 200. At
\emph{K}=100, the error-reduction factors are 33.5×, 30.5×, 25.5×,
23.8×, 23.7×, and 22.3× for \emph{N}=1024, 2048, 4096, 8192, 16384, and
65536, respectively. At K=200, the corresponding factors increase to
94.4×, 83.4×, 61.2×, 62.6×, 52.7×, and 44.9×. The improvement therefore
persists across the complete dimension range investigated.

\subsection{4.1.2 Robustness across Independent Coefficients
Vectors}\label{robustness-across-independent-coefficients-vectors}

To determine whether the observed advantage depends on a particular
coefficient-vector realization, the comparison was repeated using 50
independent coefficient vectors for every vector dimension
(\emph{N}=1024, 2048, 4096, 8192, 16384, and 65536) and the results are
shown in Table 2 by the percent of win rate. At \emph{K}=10, SETAREH-5
does not consistently outperform SETAREH-3 and its win rate ranges from
22\% to 44\%, depending on vector dimension. The behavior changes
markedly at higher retention levels. At \emph{K}=20, SETAREH-5 is better
for 98--100\% of the independent coefficient vectors, and at $K\geq 30$
it achieves a 100\% win rate for every investigated dimension.

Table 2) Robustness analysis summary on synthetic data.

\begin{longtable}[]{@{}
  >{\raggedright\arraybackslash}p{(\columnwidth - 4\tabcolsep) * \real{0.0918}}
  >{\raggedright\arraybackslash}p{(\columnwidth - 4\tabcolsep) * \real{0.3082}}
  >{\raggedright\arraybackslash}p{(\columnwidth - 4\tabcolsep) * \real{0.6001}}@{}}
\toprule\noalign{}
\begin{minipage}[b]{\linewidth}\raggedright
\emph{K}
\end{minipage} & \begin{minipage}[b]{\linewidth}\raggedright
SETAREH5 win-rate range across investigate \emph{N}
\end{minipage} & \begin{minipage}[b]{\linewidth}\raggedright
Interpretation
\end{minipage} \\
\midrule\noalign{}
\endhead
\bottomrule\noalign{}
\endlastfoot
10 & 22--44\% & Extreme-compression stress regime \\
20 & 98--100\% & Strongly consistent advantage \\
\(\geq 30\) & 100\% & Consistent across all tested dimensions \\
\end{longtable}

Because the same input-vector realizations are used for both methods for
each coefficient vector, this paired design reduces the possibility that
the observed difference is caused by different input samples. The
robustness analysis therefore supports the interpretation that the
additional transform stage of SETAREH-5 has a systematic effect at
moderate retention levels.

\subsection{\texorpdfstring{4.1.3 Accuracy and Coefficient-Level
Compression at Prescribed Error Levels
}{4.1.3 Accuracy and Coefficient-Level Compression at Prescribed Error Levels }}\label{accuracy-and-coefficient-level-compression-at-prescribed-error-levels}

A practically useful comparison is to determine how many
coefficient--input pairs are required to achieve specified approximation
accuracy. Table 3 shows the comparative results for \emph{N}=4096. As
seen, an accuracy level corresponding to a conditioned median relative
error below 5\% requires $K=30$ with SETAREH-3, whereas SETAREH-5
reaches the same target with \emph{K}=20. For a 2\% target, the required
retention decreases from \emph{K}=100 to $K=30$. For the 1\%
target, SETAREH-3 requires \emph{K}=200, whereas SETAREH-5 reaches the
target with $K=30$. Similar behavior is observed at all other
vector dimensions.

Table 3) Minimum \emph{K} required to reach the prescribed
conditioned-median relative-error targets in the synthetic baseline for
\emph{N}=4096.

\begin{longtable}[]{@{}
  >{\raggedright\arraybackslash}p{(\columnwidth - 8\tabcolsep) * \real{0.1611}}
  >{\raggedright\arraybackslash}p{(\columnwidth - 8\tabcolsep) * \real{0.1749}}
  >{\raggedright\arraybackslash}p{(\columnwidth - 8\tabcolsep) * \real{0.1719}}
  >{\raggedright\arraybackslash}p{(\columnwidth - 8\tabcolsep) * \real{0.2419}}
  >{\raggedright\arraybackslash}p{(\columnwidth - 8\tabcolsep) * \real{0.2502}}@{}}
\toprule\noalign{}
\begin{minipage}[b]{\linewidth}\raggedright
Target \emph{CMRE} (\%)
\end{minipage} & \begin{minipage}[b]{\linewidth}\raggedright
SETAREH-3 \emph{K}
\end{minipage} & \begin{minipage}[b]{\linewidth}\raggedright
SETAREH-5 \emph{K}
\end{minipage} & \begin{minipage}[b]{\linewidth}\raggedright
\emph{CR\textsubscript{3}*}
\end{minipage} & \begin{minipage}[b]{\linewidth}\raggedright
\emph{CR\textsubscript{5}*}
\end{minipage} \\
\midrule\noalign{}
\endhead
\bottomrule\noalign{}
\endlastfoot
5\% & 30 & 20 & 136.53 & 204.80 \\
2\% & 100 & 30 & 40.96 & 136.53 \\
1\% & 200 & 30 & 20.48 & 136.53 \\
\end{longtable}

Since the reduction in the required retention level directly translates
into a higher coefficient-level compression ratio, Table 3 also
demonstrates \emph{CR*} for \emph{N}=4096 at target errors for both
SETAREH variants. For example at the 1\% target, SETAREH-3 requires
\emph{K}=200, corresponding to \emph{CR\textsubscript{3}*}=20.48,
whereas SETAREH-5 requires $K=30$, corresponding to
$CR_{5}^{*}=136.5$. Thus, at this target accuracy, the
coefficient-level compression ratio is approximately 6.67 times higher
for SETAREH-5. Although it is not shown in Table 3, our study shows that
the coefficient-level compression ratio increases substantially but not
proportionally with vector dimension. For example, SETAREH-5 reaches the
approximately 1\% target with only 30 retained pairs for \emph{N}=4096,
8192, and 16384, and with 40 pairs for \emph{N}=65536. This behavior is
potentially important for high-dimensional applications because the
absolute number of retained values can remain relatively small even as
the original vector becomes substantially larger.

As a controlled numerical baseline using independently generated
Gaussian coefficient and input vectors, the first experimental setting
compared the performance of SETAREH-3 to SETAREH-5. It was noticed that
when a moderate number of transformed coefficient pairs are retained
\((K \geq 30)\), SETAREH-5 consistently outperforms SETAREH-3. In conclusion,
SETAREH-5 is chosen as the basic SETAREH algorithm to be considered in
the remaining experimental settings and for short it will be called
SETAREH rather than SETAREH-5 from now on.

\subsection{}\label{section-1}

\subsection{4.2. APPA-REAL Apparent-Age Regression: Model Construction
and SETAREH
Evaluation}\label{appa-real-apparent-age-regression-model-construction-and-setareh-evaluation}

To evaluate SETAREH in a real-world regression setting, we considered an
apparent-age estimation problem using the APPA-REAL image dataset
{[}20{]}. In this framework, each image was represented by a
4096-dimensional feature vector extracted from the fc2 layer of a
pretrained VGG16 network {[}13{]}. These feature vectors serve as the
inputs to a Ridge regression model {[}21{]} trained to estimate apparent
age.

\subsection{4.2.1. Feature Representation and Regression
Model}\label{feature-representation-and-regression-model}

The APPA-REAL dataset was divided into training, validation, and test
subsets containing 4,113, 1,500, and 1,978 images, respectively. The
feature-extraction network was used to obtain the image representations,
whereas the regression model learned the mapping from these
representations to apparent-age estimates. In continue, the Ridge
regression model development procedure is explained in brief.

Consider sample image \(i\), whose feature vector is
\(\mathbf{X}_i \in \mathbb{R}^{4096}\) and whose target apparent age is \(y_i\).
The VGG16 fc2 features were standardized using the feature-wise means
\(\mu_j\) and scales \(s_j\) estimated from the training set. The
same values were then applied unchanged to the validation and test sets:

\[Z_{ij}=\frac{X_{ij}-\mu_j}{s_j}.\tag{15}\]

Next, a Ridge regression model was trained on the standardized features
to predict apparent age using the training set. Several candidate values
of the regularization parameter \(\alpha\) were evaluated, and the value yielding
the lowest validation MAE was selected. The selected value was \(\alpha = 10000\). The model was fitted using the iterative LSQR (Least Squares QR)
algorithm, with a maximum of 100,000 iterations and a convergence
tolerance of 10\textsuperscript{-8}. The exact Ridge model achieved a
validation MAE of 8.9015 years and a validation RMSE of 11.3421 years.
On the independent test set, the corresponding MAE and RMSE were 12.0018
and 16.2052 years, respectively. These values establish the
prediction-error baseline for the trained apparent-age regression model.
In the subsequent SETAREH experiments, approximation-induced deviations
were evaluated by comparing the approximate model outputs with those of
the exact Ridge model. Thus, the MAE and RMSE reported above quantify
apparent-age prediction errors, whereas the SETAREH deviations quantify
the additional error introduced by approximating the
model\textquotesingle s inner product.

Let \(c_j\) be the fitted coefficient for standardized feature \(j\), and let \(b_{\mathrm{Ridge}}\) denote the fitted intercept. The
original pipeline predicts the apparent age as:

\[\hat{y}_i=\sum_{j=1}^{4096}c_j\frac{X_{ij}-\mu_j}{s_j}+b_{\mathrm{Ridge}}.\tag{16}\]

\subsection{4.2.2. Conversion to an Equivalent Linear
Model}\label{conversion-to-an-equivalent-linear-model}

To apply SETAREH to a single weight vector, the standardization and
Ridge regression are algebraically combined into an equivalent linear
model operating directly on the original fc2 features. Expanding
Equation (16) gives:

\[\hat{y}_i=\sum_{j=1}^{4096}\frac{c_j}{s_j}X_{ij}+b_{\mathrm{Ridge}}-\sum_{j=1}^{4096}\frac{c_j\mu_j}{s_j}.\tag{17}\]

The effective coefficients and intercept are defined as follows:

\[w_{\mathrm{eff},j}=\frac{c_j}{s_j}.\tag{18}\]

\[b_{\mathrm{eff}}=b_{\mathrm{Ridge}}-\sum_{j=1}^{4096}\frac{c_j\mu_j}{s_j}.\tag{19}\]

The equivalent prediction equation is:

\[\hat{y}_i=\mathbf{x}_i^{\mathsf T}\mathbf{w}_{\mathrm{eff}}+b_{\mathrm{eff}}.\tag{20}\]

In this experiment, \(b_{\mathrm{eff}} = 30.51889935675309\).
The transformed model was checked against the original pipeline, and their predictions agree to numerical precision. This conversion isolates the dot product \(\mathbf{x}_i^{\mathsf T}\mathbf{w}_{\mathrm{eff}}\), which is approximated by SETAREH, while the intercept remains unchanged.

\subsection{4.2.3. SETAREH Approximation
Results}\label{setareh-approximation-results}

For each image, the exact model dot product is
\(d_i = \mathbf{x}_i^{\mathsf T}\mathbf{w}_{\mathrm{eff}}\).
We applied the offline procedure of SETAREH to
\(\mathbf{w}_{\mathrm{eff}}\) and saved its transformed representation
\(\mathbf{w}_{\mathrm{eff,SETAREH}}\) and the associated permutation
information (see Figure 1). For each original fc2 feature vector
\(\mathbf{x}_i\) in the validation and test sets, the corresponding
transformed vector \(\mathbf{x}_{i,\mathrm{SETAREH}}\) is calculated
using the permutation information and the Haar discrete wavelet transform
(HDWT).

Let \(\widetilde{d}_i(K)\) denote the approximate dot product reconstructed
from the retained \(K\) coefficients by SETAREH. The approximate
apparent-age prediction is therefore:

\[\hat{y}_i(K)=\widetilde{d}_i(K)+b_{\mathrm{eff}}.\tag{21}\]

Approximation quality of SETAREH is quantified using \emph{CMRE} which
measures the approximation error introduced by coefficient truncation
compared to the original dot product. So, the maximum absolute
prediction deviation, reported in years, is the maximum difference
between approximate and exact model predictions and not the model's
age-prediction error. Table 4 summarizes the validation and test
results. \emph{CMRE} is reported as a percentage, while maximum absolute
prediction deviation is reported in years.

Table 4) SETAREH approximation results for the APPA-REAL Ridge
regression model.

\begin{longtable}[]{@{}
  >{\raggedright\arraybackslash}p{(\columnwidth - 10\tabcolsep) * \real{0.0828}}
  >{\raggedright\arraybackslash}p{(\columnwidth - 10\tabcolsep) * \real{0.1355}}
  >{\raggedright\arraybackslash}p{(\columnwidth - 10\tabcolsep) * \real{0.1980}}
  >{\raggedright\arraybackslash}p{(\columnwidth - 10\tabcolsep) * \real{0.1980}}
  >{\raggedright\arraybackslash}p{(\columnwidth - 10\tabcolsep) * \real{0.1980}}
  >{\raggedright\arraybackslash}p{(\columnwidth - 10\tabcolsep) * \real{0.1876}}@{}}
\toprule\noalign{}
\begin{minipage}[b]{\linewidth}\raggedright
\emph{K}
\end{minipage} & \begin{minipage}[b]{\linewidth}\raggedright
\emph{CR*}
\end{minipage} & \begin{minipage}[b]{\linewidth}\raggedright
\emph{CMRE\textsubscript{K}} (\%)

Validation Data Set
\end{minipage} & \begin{minipage}[b]{\linewidth}\raggedright
\emph{CMRE\textsubscript{K}} (\%)

Test Data Set
\end{minipage} & \begin{minipage}[b]{\linewidth}\raggedright
max deviation (years)

Validation Data Set
\end{minipage} & \begin{minipage}[b]{\linewidth}\raggedright
max deviation (years)

Test Data Set
\end{minipage} \\
\midrule\noalign{}
\endhead
\bottomrule\noalign{}
\endlastfoot
10 & 409.60 & 32.353 & 25.648 & 7.1011 & 7.8837 \\
20 & 204.80 & 24.696 & 20.457 & 2.6710 & 2.9005 \\
30 & 136.53 & 4.0695 & 3.4549 & 0.59094 & 0.59770 \\
40 & 102.40 & 9.7782 & 8.5652 & 1.1331 & 1.0229 \\
50 & 81.92 & 4.1442 & 3.5735 & 0.50326 & 0.42575 \\
75 & 54.61 & 0.13745 & 0.13283 & 0.047026 & 0.037877 \\
100 & 40.96 & 0.064831 & 0.060019 & 0.027701 & 0.020287 \\
150 & 27.31 & 0.020645 & 0.019567 & 0.0077254 & 0.011034 \\
200 & 20.48 & 0.0092981 & 0.0079891 & 0.0042751 & 0.0036525 \\
\end{longtable}

As seen, retaining only 75 coefficients (1.83\% of the 4096-dimensional
vector) yields a \emph{CMRE} of 0.13283\% on the test set and retaining
100 coefficients (2.44\% of the 4096-dimensional vector) reduces the
test \emph{CMRE} to 0.060019\%. At \emph{K} = 200, only 4.88\% of the
coefficients are retained, and the test \emph{CMRE} is 0.0079891\%.
Thus, a small retained subset can closely approximate the dot product
used by the original fixed regression model. It must be noted that
\emph{CMRE} is not strictly monotonic with \emph{K}. In particular, the
values of \emph{CMRE} at \emph{K} = 40 and \emph{K} = 50 exceed its
value at $K=30$. This observed behavior is consistent with the
prescribed signed-order truncation and coefficient cancellation.

Overall, APPA-REAL provides a real-feature regression evaluation of
SETAREH. It measures how accurately the method preserves the dot product
of an already-trained apparent-age regression model under coefficient
truncation.

\subsection{4.3. Amortized Storage Benefit for Multiple Input
Vectors}\label{amortized-storage-benefit-for-multiple-input-vectors}

The practical storage advantage of SETAREH becomes increasingly
significant when the same weight vector is used with multiple input
vectors. In this setting, the permutation information associated with
the fixed weight vectors is stored once and reused for all subsequent
inputs, whereas the exact representation requires storing the complete
input vector for each input instance.

For \(M\) input vectors (schematic of Figure 1), the storage required by
the dense representation can be expressed as

\begin{equation}\label{eq:dense-storage}Q_{\mathrm{dense}}(M)=MNB.\tag{22}\end{equation}

where \(N\) is the original vector length and \(B\) is the number of
bytes used to represent each coefficient. For SETAREH, the corresponding
storage can be written as

\begin{equation}\label{eq:setareh-storage}Q_{\mathrm{SETAREH}}(M)=Q_P+MK(B+B_I).\tag{23}\end{equation}

where \(Q_{P}\) denotes the one-time storage cost of the required
permutation information, \(K\) is the number of retained
transform-domain coefficients per input, and \(B_{I}\) is the number of
bytes used for each coefficient index. Consequently, the amortized
storage per input vector is

\begin{equation}\label{eq:amortized-storage}\overline{Q}_{\mathrm{SETAREH}}(M)=\frac{Q_P}{M}+K(B+B_I).\tag{24}\end{equation}

This expression highlights an important characteristic of SETAREH: the
contribution of the fixed permutation metadata decreases inversely with
the number of input vectors. Thus, although the permutation overhead can
substantially reduce the benefit for a single input, its relative
contribution becomes progressively smaller as the same weight
representation is reused.

For the APPA-REAL ridge regression configuration considered in this
study (\(N = 4096\)), using 4-byte coefficients and 2-byte indices, the
three stored permutations require approximately 24 KiB. For \(K = 75\),
the SETAREH representation requires approximately 0.43945 KiB per
additional input vector. Therefore, the total storage is

\begin{equation}\label{eq:setareh-storage-numeric}Q_{\mathrm{SETAREH}}(M)=24+0.43945M\quad(\mathrm{KiB}).\tag{25}\end{equation}

whereas the corresponding dense representation requires

\begin{equation}\label{eq:dense-storage-numeric}Q_{\mathrm{dense}}(M)=16M\quad(\mathrm{KiB}).\tag{26}\end{equation}

Therefore, the percent of memory saving by SETAREH for \emph{M} input
feature vectors compared to the original ones are:

\begin{equation}\label{eq:saving}\mathrm{Saving}(M)=\frac{Q_{\mathrm{dense}}(M)-Q_{\mathrm{SETAREH}}(M)}{Q_{\mathrm{dense}}(M)}\times100\%.\tag{27}\end{equation}

Table 5 compares memory needed for dense representations and SETAREH
representations for different values of input vectors \emph{M}.

Table 5) The memory needed (in KiB) for dense and SETAREH
representations for different number of input vectors

\begin{longtable}[]{@{}
  >{\raggedright\arraybackslash}p{(\columnwidth - 6\tabcolsep) * \real{0.2881}}
  >{\raggedright\arraybackslash}p{(\columnwidth - 6\tabcolsep) * \real{0.2511}}
  >{\raggedright\arraybackslash}p{(\columnwidth - 6\tabcolsep) * \real{0.3043}}
  >{\raggedright\arraybackslash}p{(\columnwidth - 6\tabcolsep) * \real{0.1565}}@{}}
\toprule\noalign{}
\begin{minipage}[b]{\linewidth}\raggedright
Number of input vectors

\emph{M}
\end{minipage} & \begin{minipage}[b]{\linewidth}\raggedright
Dense representation

(KiB)
\end{minipage} & \begin{minipage}[b]{\linewidth}\raggedright
SETAREH representation

(KiB)
\end{minipage} & \begin{minipage}[b]{\linewidth}\raggedright
\[Saving(M)\]
\end{minipage} \\
\midrule\noalign{}
\endhead
\bottomrule\noalign{}
\endlastfoot
1 & 16.00 & 24.44 & −52.7\% \\
2 & 32.00 & 24.88 & 22.3\% \\
10 & 160.00 & 28.39 & 82.3\% \\
100 & 1600.00 & 67.95 & 95.8\% \\
1000 & 16000.00 & 463.45 & 97.1\% \\
\end{longtable}

As seen, the resulting amortization is substantial. For a single input
vector (\(M = 1\)), SETAREH is larger than the dense representation
because of the one-time permutation overhead. However, the
representation becomes smaller from the second input onward and the
saving increases rapidly with \(M\). For example, at \(M = 2\),
\(M = 10\), \(M = 100\), and \(M = 1000\), the storage savings are
approximately 22.3\%, 82.3\%, 95.8\%, and 97.1\%, respectively.

These results demonstrate that SETAREH is particularly well suited to
applications in which a fixed set of weights is repeatedly applied to
many input vectors. In such scenarios, the one-time cost of storing the
permutations is amortized over the input set, allowing the retained
coefficient representation to dominate the overall storage cost. This
amortization effect is therefore an important practical advantage of
SETAREH beyond the coefficient-level reduction reported in the preceding
experiments.

\section{5. Conclusions}\label{conclusions}

In this paper a novel Haar-wavelet transform based sparsification
approach for reducing the number of input-vector samples required to
approximate an inner product with a fixed high-dimensional coefficient
vector was presented. Such inner-product operations arise in many
neural-network computations, including fully connected layers and, more
generally, linear projections used in convolutional and attention-based
architectures. This makes the proposed method potentially relevant to
systems in which the same parameters are repeatedly applied to a large
number of input vectors.

We called the proposed algorithm SETAREH (Sorting-assisted
Energy-concentrating Transform for Approximate Representation using Haar
wavelets) where the central idea is based on the permutation invariance
of the inner product which allows to reorder the coefficients vector so
that, after Haar-wavelet transformation, its representation can be
approximated using a relatively small number of transformed
coefficients--input pairs. This enables aggressive truncation while
retaining the potential to preserve the inner-product value with
relatively small approximation error, enabling reducing the amount of
input data that must be stored, transferred, or processed, providing a
complementary form of data reduction without modifying the underlying
network architecture or its learned parameters.

Two variants of the proposed method, namely SETAREH-3 and SETAREH-5,
were investigated. The synthetic experiments showed that SETAREH-5
generally achieved lower CMRE than SETAREH-3 for \(K\geq 20\), with a
consistent advantage at \(K\geq 30\) across the tested dimensions. Therefore,
it was selected as the main SETAREH algorithm.

To evaluate the performance of SETAREH at a real scenario, we considered
an apparent-age estimation problem using the APPA-REAL image dataset
where each input image was represented by a 4096-dimensional feature
vector extracted from the fc2 layer of a pretrained VGG16 network and
each vector serves as the input to a Ridge regression model trained to
estimate apparent age.

The study showed that by retaining only 75 processed coefficients
compared to the 4096 original coefficients (1.83\%) yielded a
\emph{CMRE} of 0.13283\% on the test set and retaining 200 coefficients
(4.88\%) the test \emph{CMRE} is 0.0079891\%. Thus, a small retained
subset can closely approximate the dot product used by the original
fixed regression model.

An important distinction should nevertheless be made between
coefficient-level reduction and complete representation compression. The
coefficient-only compression ratio \emph{N}/\emph{K} reported in this
work describes the reduction in the number of retained
coefficient--input values. It does not include the storage or
transmission cost of the permutation information required by the current
implementation. Therefore, the reported ratios should not be interpreted
as direct end-to-end model-storage compression ratios. This distinction
is important when comparing SETAREH with established model-compression
techniques such as pruning, quantization, and entropy coding
{[}7,9,15,16{]}. The conceptual position of SETAREH is therefore
complementary to conventional neural-network compression {[}14{]}.

The storage analysis shows that the practical benefit of SETAREH depends
on the number of input vectors processed using the same fixed
coefficient vector. Under the storage assumptions described in Section
4.3, SETAREH requires more storage than the dense representation for a
single input vector because of the one-time permutation overhead,
resulting in a storage increase of approximately 52.7\%. From the second
input vector onward, the SETAREH representation becomes smaller than the
dense representation. The resulting storage savings are approximately
22.3\%, 82.3\%, 95.8\%, and 97.1\% for 2, 10, 100, and 1000 input
vectors, respectively. These results illustrate the importance of
amortizing the permutation metadata across repeated inputs.

The results should also not be interpreted as evidence that SETAREH
necessarily reduces arithmetic computation in its current
implementation. The sorting and Haar-transform stages introduce
additional operations, and the present study primarily evaluates
approximation accuracy and coefficient-level data reduction rather than
end-to-end hardware latency or energy consumption. Thus, the principal
contribution demonstrated by the present experiments is structured
reduction of the number of retained coefficient--input pairs, rather
than demonstrated computational acceleration {[}17{]}.

The transform-domain formulation of SETAREH may be relevant beyond
neural networks. Any computational system containing repeated inner
products between a fixed or slowly changing coefficient vector and a
large number of input vectors could potentially benefit from the
proposed representation. Candidate areas include signal processing,
similarity computation, embedded sensing, and selected scientific or
biomedical computing workloads {[}19{]}. These applications, however,
require dedicated validation and should be considered potential
directions rather than demonstrated applications of the present study.

The characteristics demonstrated by SETAREH suggest several potential
application areas including communication-constrained neural-network
inference for edge devices, or distributed computing nodes {[}8,16{]},
memory-constrained inference, edge and embedded inference and
distributed or federated machine learning {[}18{]}, where communication
cost can dominate system performance.

Future work could develop an efficient representation of the permutation
metadata and evaluate the true end-to-end storage and communication cost
of SETAREH. This would allow coefficient-level reduction to be converted
into a rigorous system-level compression metric.

Moreover, hardware-aware implementations should be investigated to
quantify latency, memory traffic, energy consumption, and the trade-off
between transformation overhead and data-reduction benefits. Such
studies would determine whether SETAREH can provide a practical
advantage in real edge, accelerator, or distributed-computing systems.

\textbf{Data Availability.}

The APPA-REAL dataset is publicly available at the ChaLearn LAP website:
\url{https://chalearnlap.cvc.uab.es/dataset/26/description/}. The
dataset includes facial images annotated with real and apparent age
labels. Further details are provided in Agustsson et al. {[}20{]}.

\section{}\label{section-2}

\section{Acknowledgment}\label{acknowledgment}

The author acknowledges the use of OpenAI\textquotesingle s ChatGPT as
an AI-assisted tool for MATLAB code development, data analysis, and
language editing. The author is solely responsible for the scientific
content, results, interpretations, and final manuscript.

\section{}\label{section-3}

\section{References}\label{references}

{[}1{]} T. B. Brown et al., ``Language Models are Few-Shot Learners,''
Advances in Neural Information Processing Systems (NeurIPS), 2020.

{[}2{]} W. Fedus, B. Zoph, and N. Shazeer, ``Switch Transformers:
Scaling to Trillion Parameter Models with Simple and Efficient
Sparsity,'' Journal of Machine Learning Research, vol. 23, no. 120, pp.
1--39, 2022

{[}3{]} Y. Gusak et al., ``Survey on Efficient Training of Large Neural
Networks,'' IJCAI, 2022, pp. 5494--5501 ; DOI: 10.24963/ijcai.2022/769

{[}4{]} K. He, X. Zhang, S. Ren, and J. Sun, ``Deep Residual Learning
for Image Recognition,'' Proceedings of the IEEE Conference on Computer
Vision and Pattern Recognition (CVPR), pp. 770--778, 2016.

{[}5{]} A. G. Howard et al., ``MobileNets: Efficient Convolutional
Neural Networks for Mobile Vision Applications,'' arXiv:1704.04861,
2017.

{[}6{]} M. Tan and Q. V. Le, ``EfficientNet: Rethinking Model Scaling
for Convolutional Neural Networks,'' Proceedings of the 36th
International Conference on Machine Learning, PMLR 97, pp. 6105--6114,
2019.

{[}7{]} S. Han, H. Mao, and W. J. Dally, ``Deep Compression: Compressing
Deep Neural Networks with Pruning, Trained Quantization and Huffman
Coding,'' International Conference on Learning Representations (ICLR),
2016. doi:10.48550/arXiv.1510.00149.

{[}8{]} Y.-H. Chen, J. Emer, and V. Sze, ``Eyeriss: A Spatial
Architecture for Energy-Efficient Dataflow for Convolutional Neural
Networks,'' Proceedings of the 43rd Annual International Symposium on
Computer Architecture (ISCA), pp. 367--379, 2016.

{[}9{]} B. Jacob et al., ``Quantization and Training of Neural Networks
for Efficient Integer-Arithmetic-Only Inference,'' Proceedings of the
IEEE/CVF Conference on Computer Vision and Pattern Recognition (CVPR),
pp. 2704--2713, 2018.

{[}10{]} A. Haar, ``Zur Theorie der orthogonalen Funktionensysteme,''
Mathematische Annalen, vol. 69, pp. 331--371, 1910.
doi:10.1007/BF01456326.

{[}11{]} C. K. Chui, An Introduction to Wavelets, Academic Press, San
Diego, 1992.

{[}12{]} S. G. Mallat, ``A Theory for Multiresolution Signal
Decomposition: The Wavelet Representation,'' IEEE Transactions on
Pattern Analysis and Machine Intelligence, vol. 11, no. 7, pp. 674--693,
1989. doi:10.1109/34.192463.

{[}13{]} K. Simonyan and A. Zisserman, ``Very deep convolutional
networks for large-scale image recognition,'' in \emph{Proc. 3rd Int.
Conf. Learn. Representations (ICLR)}, San Diego, CA, USA, 2015.

{[}14{]} M. Wolter, S. Lin, and A. Yao, ``Neural Network Compression via
Learnable Wavelet Transforms,'' Lecture Notes in Computer Science, 2020.
doi:10.1007/978-3-030-61616-8\_4.

{[}15{]} S. Han, J. Pool, J. Tran, and W. J. Dally, ``Learning both
Weights and Connections for Efficient Neural Network,'' Advances in
Neural Information Processing Systems, 2015.

{[}16{]} L. Deng et al., ``Model Compression and Hardware Acceleration
for Neural Networks: A Comprehensive Survey,'' Proceedings of the IEEE,
vol. 108, no. 4, pp. 485--532, 2020, doi: 10.1109/JPROC.2020.2976475

{[}17{]} Q. Zhang et al., ``Recent advances in convolutional neural
network acceleration,'' Neurocomputing, vol. 323, pp. 37--51, 2019, DOI
10.1016/j.neucom.2018.09.038

{[}18{]} N. He et al., ``Communication-Efficient Adaptive
Wavelet-Compressed Federated Learning for State of Health Prediction of
Lithium-Ion Batteries in IoT Environments,'' IEEE Internet of Things
Journal, 2025.

{[}19{]} F. Dai, ``Deep learning based medical image compression using
cross attention learning and wavelet transform,'' Scientific Reports,
vol. 15, 40008, 2025. doi:10.1038/s41598-025-23582-y.

{[}20{]} E. Agustsson, R. Timofte, S. Escalera, X. Baró, I. Guyon, and
R. Rothe, ``Apparent and real age estimation in still images with deep
residual regressors on APPA-REAL database,'' in Proc. 12th IEEE Int.
Conf. Automatic Face \& Gesture Recognition (FG 2017), Washington, DC,
USA, 2017, pp. 87--94, doi: 10.1109/FG.2017.20.

{[}21{]} A. E. Hoerl and R. W. Kennard, ``Ridge regression: Biased
estimation for nonorthogonal problems,'' \emph{Technometrics}, vol. 12,
no. 1, pp. 55--67, Feb. 1970, doi: 10.1080/00401706.1970.10488634.

\end{document}